\documentclass[11pt,a4paper]{article}
\usepackage[hyperref]{acl2021}
\usepackage{times}
\usepackage{latexsym}

\usepackage{microtype}

\aclfinalcopy 

\usepackage{booktabs}
\usepackage{ltablex}
\usepackage{xspace}
\usepackage{url}
\usepackage{arydshln}
\usepackage{multirow}
\usepackage{adjustbox}

\newcommand{\model}[1]{{#1}\xspace}
\newcommand{\electrabase}{\model{ELECTRA}}
\newcommand{\robertabase}{\model{RoBERTa}}

\newcommand{\mobilebert}{\model{MobileBERT}}
\newcommand{\squeezebert}{\model{SqueezeBERT}}

\newcommand{\electrabasemrc}{\electrabase{}(T)\xspace}
\newcommand{\robertabasemrc}{\robertabase{}(T)\xspace}

\newcommand{\mobilebertmrc}{\mobilebert{}(T)\xspace}
\newcommand{\squeezebertmrc}{\squeezebert{}(T)\xspace}

\newcommand{\tabref}[2][]{Table#1~\ref{table:#2}}

\title{Does QA-based intermediate training help fine-tuning language models for text classification?}

\author{Shiwei Zhang\\
  School of Computing Technologies\\
  RMIT University, Australia\\
  \texttt{dr.shiwei.zhang@gmail.com} \\\And
  Xiuzhen Zhang\Thanks{Corresponding author.} \\
  School of Computing Technologies\\
  RMIT University, Australia \\
  \texttt{xiuzhen.zhang@rmit.edu.au} \\}

\date{}

\begin{document}
\maketitle
\begin{abstract}
Fine-tuning pre-trained language models for downstream tasks has become a norm for NLP.  
Recently it is found that intermediate training based on high-level inference tasks such as Question Answering (QA) can improve the performance of some language models for target tasks. 
However it is not clear if intermediate training generally benefits various language models.  
In this paper, using the SQuAD-2.0 QA task for intermediate training for target text classification tasks,
we experimented on eight tasks for single-sequence classification and eight tasks for sequence-pair classification using two base and two compact language models. 
Our experiments show that QA-based intermediate training 
generates varying transfer performance across different language models, except for similar QA tasks. 
\end{abstract}

\section{Introduction}

The framework of fine-tuning pre-trained Language models (LMs), especially transformer-based LMs, for downstream tasks has shown state-of-the-art performance on many natural language processing (NLP) tasks~\cite{devlin2019bert,raffel2020exploring}.
It is believed that the pre-training stage leads LMs to develop general-purpose abilities and knowledge that can then be transferred to downstream tasks~\cite{raffel2020exploring}.

To further improve the performance of pre-trained LMs on target tasks, two novel training approaches have been recently researched, namely further pre-training and intermediate training.
A further pre-training stage for LMs~\cite{gururangan2020don} is a stage between pre-training and fine-tuning, which further pre-trains LMs on an extra dataset using unsupervised objectives. 
It has been found that further pre-training LM on the target domain (domain-adaptive pre-training) leads to improved performance on target tasks~\cite{gururangan2020don}.
Another effective transfer learning approach named intermediate training that chooses to train a LM model on an intermediate task via supervised manner and then fine-tune it on target tasks.
This also leads to promising results across various NLP tasks including text classification, QA and sequence labeling~\cite{phang2018sentence,vu2020exploring,pruksachatkun2020intermediate}.

\begin{figure}[t]
	\centering
	\includegraphics[width=0.9\linewidth]{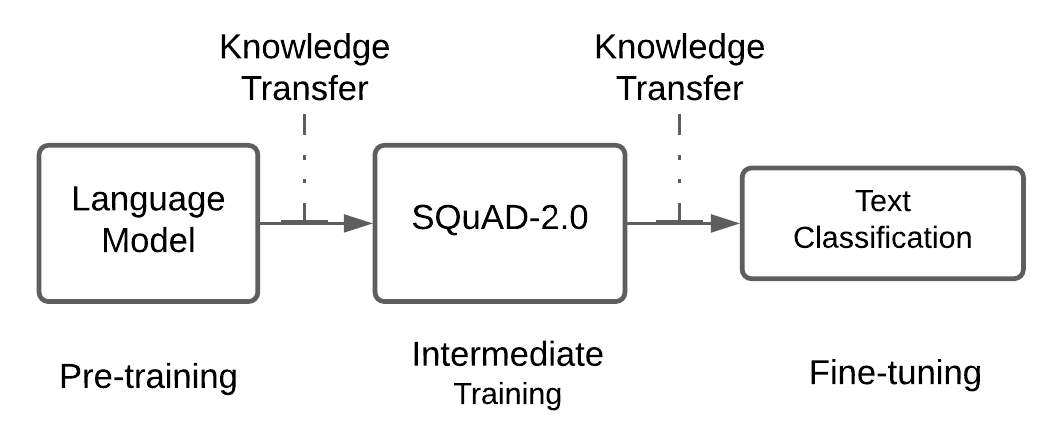}
	\caption{We experiment SQuAD-2.0 as the intermediate training task for text classification tasks.}
	\label{fig:framework}
\end{figure}

Text classification is the problem of classifying text into categories or classes which has been widely studied.
In terms of input, there are mainly two forms of text classification problems: single-sequence classification tasks (e.g., sentiment classification and topic classification) and pairwise tasks (e.g., NLI and IR-related QA).
In recent years, a common approach to tackle text classification problems is to fine-tune a pre-trained LM on target text classification tasks.
Recently, advanced transfer learning-based approaches have been proposed to further improve the performance.
For example, a recent work~\cite{sun2019fine} has studied how to fine-tune BERT for text classification.
They found that further pre-training LM using data within-task or in-domian can improve the performance of BERT for text classification tasks.

More recently, cross-task transfer learning technique for text classification has been investigated~\cite{vu2020exploring}, and it is
found that tasks that require high-level inference and reasoning abilities, such as natural language inference and question answering (QA)~\cite{rajpurkar2018know}, are often the best intermediate tasks for text classification tasks.
In a recent study~\cite{pruksachatkun2020intermediate}, it is found that natural language inference and QA tasks are generally helpful as intermediate tasks.
\citet{vu2020exploring} showed that SQuAD-2.0 is the most favourable intermediate task for text classification. 
There are only a few text classifications tasks and only one language model (BERT) in their experiments,
making it hard to conclude that SQuAD-2.0 as the intermediate task can generally improve the performance of all types of text classification tasks.

In this paper, we investigate the effectiveness of intermediate training for four different LMs -- \electrabase, \robertabase, \mobilebert, and \squeezebert)-- 
using the most popular QA resource SQuAD-2.0 as the intermediate task for eight target text classification tasks.
We found that intermediate training shows varying transfer performance across different language models.
Particularly contrary to previous thoughts, intermediate training with high-level inference QA tasks does not generally show positive transfer for low-level inference text classification tasks. 

\section{Related Work}
As a large quantity of labeled data is not always available for training deep learning models, transfer learning becomes quite important for many of NLP problems.
With transfer learning, widely available unlabeled text corpora containing rich semantic and syntactic information can be leveraged for learning language models, such as BERT~\cite{devlin2019bert}, GPT~\cite{brown2020language}, and T5~\cite{raffel2020exploring}.
Then, these language models are fine-tuned on downstream tasks, which is the dominant transfer learning method adopted in NLP at the moment.
The second way of using transfer learning in NLP is to further pre-train pre-trained language models in domain data before fine-tuning on downstream tasks~\cite{gururangan2020don,sun2019fine}. 
The third approach, which is the method we investigate in our work, is to transfer models fine-tuned on an intermediate task for a target task~\cite{pruksachatkun2020intermediate}.

A recent work~\cite{pruksachatkun2020intermediate} investigated when and why intermediate-task training is beneficial for a given target task.
They experimented with 11 intermediate tasks and 10 target tasks, and find that intermediate tasks requiring high-level inference and reasoning abilities tend to work best, such as natural language inference tasks and QA tasks.
Another recent work~\cite{vu2020exploring} has explored transferability across three types of tasks, namely text classification/regression, question answering and sequence labeling.
They found that transfer learning is more beneficial for low-data source tasks and also found that data size, task and domain similarity, and task complexity all can affect transferability.


\begin{table*}[t]
	\begin{center}
		\caption{Dataset Statistics}
		\label{table:dataset}
		\begin{adjustbox}{max width=\linewidth}
			\begin{tabular}{lrrlrr}
				\toprule
				&Task &\#DataSize (Training/Testing)  &\#Classes &Metric & Source  \\ \midrule 
				{AGNEWS}~\cite{zhang2015character} & Topic Classification & 120000/7600 & {0: 31900, 1: 31900, 2: 31900, 3: 31900} & Accuracy &News\\
				{SST2}~\cite{wang2018glue} & Sentiment Classification & 67349/872 & {0: 30208, 1: 38013} & Accuracy & Movie Reviews\\
				{LIAR}~\cite{wang2017liar} & Fake News Detection & 10269/1283 & {0: 2248, 1: 2390, 2: 2215, 3: 1894, 4: 1871, 5: 934} & F1 &POLITIFACT.COM\\ 
				{OFFENSIVE}~\cite{barbieri2020tweeteval} & Offensive Speech Detection & 11916/1324 & {0: 8595, 1: 4181} & F1 &Twitter \\ 
				{HATE}~\cite{barbieri2020tweeteval} & Hate Speech Detection & 9000/2970 & {0: 6935, 1: 5035} & F1 &Twitter \\
				{COLA}~\cite{wang2018glue} & Linguistic Acceptability & 8551/1043 & {0: 2850, 1: 6744} & Matthews Correlation & Books and Journal \\ 
				{EMOTION}~\cite{barbieri2020tweeteval} & Emotion Detection & 3257/1421 & {0: 1958, 1: 1066, 2: 417, 3: 1237} & F1 &Twitter \\
				{IRONY}~\cite{barbieri2020tweeteval} & Irony Detection & 2862/784 & {0: 1890, 1: 1756} & F1 &Twitter \\ \hline
				
				{MNLI}~\cite{wang2018glue} & Natural Language Inference & 392702/9815 & {0: 134378, 1: 134023, 2: 134116}  & Accuracy &Multiple Text Corpus\\
				{QQP}~\cite{wang2018glue}  & Quora Question Pairs & 363846/40430 & {0: 255013, 1: 149263} & F1 & Quora \\
				{QNLI}~\cite{wang2018glue} & Question Answering & 104743/5463 & {0: 55079, 1: 55127} & Accuracy &  Wikipedia \\
				{WIKIQA}~\cite{yang2015wikiqa} & Question Answering & 20360/2733 & {0: 25192, 1: 1333} & F1 & Wikipedia \\
				{BOOLQ}~\cite{wang2019superglue} & Boolean Questions & 9427/3270 & {0: 4790, 1: 7907} & F1 & Google search \\
				{MRPC}~\cite{wang2018glue} & Semantic Equivalence & 3668/408 &{0: 1323, 1: 2753} & F1 & News \\ 
				{RTE}~\cite{wang2018glue} & Recognizing Textual Entailment  & 2490/277 & {0: 1395, 1: 1372} & Accuracy & News and Wikipedia \\
				{WNLI}~\cite{wang2018glue} &  Natural Language Inference & 635/71 & {0: 363, 1: 343} & Accuracy &  Winograd Schema Challenge\\ 
				\bottomrule
			\end{tabular}
		\end{adjustbox}
	\end{center}
\end{table*}

\section{Methods}
To find out whether using SQuAD-2.0 as the intermediate training task is generally helpful for text classification tasks for different language models, we experiment with 8 single-sequence text classification tasks and 8 sequence-pair text classification tasks, across four language models.

In SQuAD-2.0, each question is given a context from which to infer the answer. 
A QA system is expected to extract a span of text from that given context.
More specifically, given a context $C$ which consists of $n$ tokens ($[t_1,t_2,...t_n]$) and a question $Q$, a QA model is expected to predict the position of the start and end tokens of the answer in the context $C$. 
To correctly extract the answer span, on one hand an SQuAD-2.0 model needs to learn word-level dependencies between two sequences (semantic similarity); on the other hand it learns how to infer an answer from the context given a question.
Training a transformer-based LM for SQuAD-2.0 intuitively enforces model's ability on inference and measuring semantic similarity, which 
is shown in previous studies~\cite{pruksachatkun2020intermediate,vu2020exploring}
to benefit text classification target tasks at the lower, sequence-level, 
either classification of  single sequences or classification of the inference or similarity for sequence pairs.

When using transformer-based models for pairwise text classification, often a special token (e.g., [SEP]) is added between two sequences, similar to the QA input.
We are interested in whether such a similarity between QA tasks and sequence-pair text classification tasks can make a difference.
In terms of training procedure, we follow previous works~\cite{phang2018sentence,vu2020exploring}.
Specifically, we first fine-tune a pre-trained LM on SQuAD-2.0 (intermediate training stage) and then fine-tune it on each text classification tasks.

When adopting transformer-based language models (LM) for span extraction, we first load a pre-trained LM and then add a span classification head on top of it (a linear layer on top of the hidden-states output).
A span classification head eventually generates two logits for each token, namely a logit for the start token and a logit for the end token.
Learning a SQuAD-2.0 model performs classification at the token-level -- classify a token either the start token or the end token.
At inference stage,  predictions are made based on logits (taking the token with the largest start logits as a start token and the token with largest end logits as an end token). 

After we train a SQuAD-2.0 model, the next step is to transfer it for text classification tasks.
When transferring a SQuAD-2.0 model, we only need to change a span classification head to a sequence classification head.
The transferred transformer with a new sequence classification head will then be fine-tuned on text classification tasks.
The weights of both the transferred SQuAD-2.0 model and the classification head will be updated during the fine-tuning stage.
Therefore, the training process consists of three training stages, namely pre-training stage (pre-training a LM), intermediate training stage (fine-tuning on SQuAD-2.0), and fine-tuning stage (fine-tuning on each text classification tasks).

\section{Experiments}

\subsection{Data and models}
The dataset statistics and evaluation metrics for each task are shown in \tabref{dataset}.
We selected 8 single-sequence text classification tasks and 8 sequence-pair text classification tasks,
covering binary and multi-class classification problems, balanced and imbalanced datasets, data-rich and data-scarce tasks, and different data sources.
We select four pre-trained transformer-based LMs, namely ELECTRA~\cite{clark2019electra}, RoBERTa~\cite{liu2019roberta}, MobileBERT~\cite{sun2020mobilebert}, SqueezeBERT~\cite{iandola2020squeezebert}.

 \begin{table*}[t]
	\centering 
	\begin{tabular}{lllllllll}\toprule
		& \small{AGNEWS} & \small{SST2} & \small{LIAR} & \small{OFFENSIVE} &  \small{HATE} & \small{COLA} & \small{EMOTION}  & \small{IRONY}  \\ \hline 
		
		\electrabase         & 94.46 & 94.61 & 26.63  & 83.48 & 48.01 & 67.65 & 82.59  & 71.96 \\ 
		\electrabasemrc      & 94.59$^+$ & 94.26$^-$ & 27.76$^+$ & 82.91$^-$  & 44.90$^-$ & 67.01$^-$ & 81.86$^-$  & 70.96$^-$  \\  \hline
		
		\robertabase         & 94.84 & 93.00 & 27.65 & 83.18 & 44.19   & 58.84 & 82.75 & 71.41 \\     
		\robertabasemrc      & 94.82$^=$ & 94.15$^+$ & 27.35$^-$ & 83.45$^+$ & 46.62$^+$   & 57.17$^-$ & 81.79$^-$ & 69.35$^-$ \\ \hline
		
		\mobilebert      & 94.57 & 90.13 & 26.07  & 84.71 & 43.66 & 49.99 & 78.23 & 63.08  \\
		\mobilebertmrc   & 94.32$^-$ & 91.05$^+$ & 26.27$^+$  & 85.01$^+$ & 45.57$^+$ & 50.25$^+$ & 79.72$^+$  & 62.36$^-$ \\ \hline
		
		\squeezebert    & 94.68 & 89.90 & 27.26  & 84.09 & 41.97 & 44.50 & 78.72  & 66.07 \\ 
		\squeezebertmrc & 94.09$^-$ & 89.10$^-$ & 27.72$^+$ & 83.61$^-$ & 40.54$^-$ & 35.37$^-$ & 77.73$^-$  & 66.44$^+$  \\
		
		\bottomrule
	\end{tabular}
	\caption{Performance(\%) for single-sequence text classification tasks. Models with SQuAD2.0 intermediate tuning are denoted with T, $+$, $=$ and $-$ denote increase, equal and decrease in performance for SQuAD-tuned models.}
	\label{table:results01}
\end{table*}

%
%
		
		
		

\begin{table*}
	\centering 
	\begin{tabular}{lllllllll}\toprule
		& \small{QQP} & \small{\bf QNLI} & \small{WNLI} &  \small{MNLI} & \small{\bf WIKIQA} & \small{\bf BOOLQ} & \small{MRPC} & \small{RTE} \\ \hline 
		
		\electrabase         & 91.69 & 92.09 & 47.88 & 88.52 & 46.04 & 84.16 & 88.60 & 77.61\\ 
		\electrabasemrc      & 91.45$^-$ & {\bf 92.44$^+$} & 52.58$^+$ & 88.77$^+$ & {\bf 50.43$^+$} & {\bf 86.34$^+$} & 87.78$^-$ & 78.34$^+$ \\  \hline
		
		\robertabase         & 91.24 & 92.04 & 56.34 & 87.69 & 43.41 & 84.22 & 89.56 & 75.33 \\     
		\robertabasemrc      & 91.14$^-$ & {\bf 92.42$^+$} & 56.34$^=$ & 87.65$^=$ & {\bf 52.45$^+$} & {\bf 84.54$^+$} & 88.31$^-$ & 79.18$^+$ \\ \hline
		
		\mobilebert        & 89.09 & 89.18 & 46.48 & 82.63 & 40.18 &77.65 & 83.69 & 56.68 \\
		\mobilebertmrc    & 88.94$^-$ & {\bf 90.88$^+$} & 35.21$^-$ & 82.45$^-$ & {\bf 52.60$^+$} & {\bf 81.63$^+$} & 86.87$^+$ & 67.75$^+$ \\ \hline
		
		\squeezebert    & 89.32 & 89.16 & 52.11 & 80.49 & 41.70 & 79.45 & 83.62 & 68.11 \\ 
		\squeezebertmrc  & 89.07$^-$ & {\bf 90.13$^+$} & 39.90$^-$ & 80.05$^-$ & {\bf 50.89$^+$} & {\bf 79.98$^+$} & 85.31$^+$ & 66.79$^-$ \\
		
		\bottomrule
	\end{tabular}
	\caption{Performance(\%) for pairwise classification tasks. Models with SQuAD2.0 intermediate tuning are denoted with T, where $+$, $=$ and $-$ denote increase, equal and decrease in performance for SQuAD-tuned models. Note the positive transfer results on QA tasks QNLI, WIKIQA and BOOLQ.}
	\label{table:results02}
\end{table*}

\subsection{Results}

Experiment results (averaged over three runs) are reported in \tabref{results01} and \tabref{results02}.
Note that QQP, QNLI, MNLI, MRPC, WNLI, RTE, and COLA are sub-tasks of language understanding benchmark GLUE~\cite{wang2018glue} widely used for LM evaluation.
Our results are slightly different from (lower than) those reported in their paper, as we used the same setting of hyper-parameters (e.g., epoch, learning rate, input length, and batch size) for all LMs rather than tuning hyper-parameters, for fair comparison across all LMs.

According to \tabref{results01}, we can see that SQuAD2-tuned models for single-sequence text classification tasks have mixed results.
On data-rich tasks, such as AGNEWS and SST2, the performance of SQuAD2-tuned models are slightly worse, except for \robertabasemrc and \mobilebertmrc which have slightly better performance on SST2.
On data-poor tasks, such as IRONY and EMOTION, transferred SQuAD2 models also tend to perform worse.
In case of multi-class problems, such as AGNEWS and LIAR, the performance of models with SQuAD2 fine-tuning are not consistent.
For example, \electrabasemrc,  \mobilebertmrc and \squeezebertmrc improved the performance on LIAR, while \robertabasemrc did not.
Overall, we can see that SQuAD2-tuned models show varying transfer performance across four language models for single-sequence classification.

The results of sequence-pair text classification are reported in \tabref{results02}.
Sequence-pair tasks can be roughly categorized into two groups, namely similarity tasks (e.g., QQP, MPRC) and inference tasks.
Similarity tasks measure the semantic similarity between two sequences, while inference tasks measure the semantic relations between two sequences.
Inference tasks have two sub-groups: natural language inference (e.g., WNLI, MNLI and RTE) and QA-related tasks (e.g., QNLI, WIKIQA and BOOLQ). 
We can see that SQuAD2-tuned models have consistently better performance for QA tasks 
QNLI, WIKIQA and BOOLQ.
A possible explanation is when trained on SQuAD-2.0, if a question is unanswerable, the index of [CLS] token is usually set as the answer, which means that the representation of [CLS] token contains information about whether a question has the answer in the given context. 
On similarity tasks, SQuAD2-tuned models have worse performance on QQP (data-rich), but on MRPC (data-poor) SQuAD2-tuned models tend to have mixed performance.
On natural language inference tasks, MNLI (data-rich) seems not benefit from SQuAD2 fine-tuning, but the performance on WNLI (data-poor) has shown some improvements. 
Our experiments show that SQuAD2-tuned models have seen consistent success on QA tasks, but generally
sequence-pair tasks do not always benefit from this intermediate training, whether data rich or data-poor.
Consequently, it is still hard to conclude that using SQuAD-2.0 as the intermediate training task is generally helpful for text classification. 

\section{Conclusion}
We studied using the SQuAD-2.0 QA intermediate task for target text classification across different language models.
Our experiments on eight classification target tasks and four language models show that SQuAD2-tuned models do not generally have better performance, whether single-sequence or sequence-pair, or data-rich or data-poor settings.
This result highlights that high-level inference intermediate tasks may not generally produce positive transfer as previously thought. 
%
On the other hand, SQuAD-tuned models always have 
positive transfer results for QA tasks, 
which suggests further research is needed to investigate if  task similarity rather than task complexity plays a significant role for intermediate training.

\section*{Acknowledgements}
This initiative was funded by the Australian government Department of Defence and the Office of National Intelligence under the AI for Decision Making Program, delivered in partnership with the Defence Science Institute in Victoria. 

\bibliographystyle{acl_natbib}
\bibliography{anthology,acl2021}


\end{document}